%% file: head.tex
\documentclass[sigconf]{acmart}
\renewcommand\footnotetextcopyrightpermission[1]{}

\usepackage{subfigure}
\usepackage{mathrsfs}
\usepackage{algpseudocode}
\usepackage{algorithmicx}
\usepackage[ruled]{algorithm2e}
\usepackage[utf8]{inputenc}
\usepackage{array}
\usepackage{booktabs}
\usepackage{multirow}
\usepackage{graphicx}
\usepackage{enumitem}
\usepackage{verbatimbox}
\setcopyright{acmcopyright}
\def\BibTeX{{\rm B\kern-.05em{\sc i\kern-.025em b}\kern-.08emT\kern-.1667em\lower.7ex\hbox{E}\kern-.125emX}}
\usepackage{soul}

\begin{document}
\title{Complexity-aware Large Scale Origin-Destination Network Generation via Diffusion Model}

\author{Can Rong}
\affiliation{%
  \institution{Tsinghua University}
  \city{Beijing}
  \country{China}
}
\email{rc20@mails.tsinghua.edu.cn}

\author{Jingtao Ding}
\affiliation{%
  \institution{Tsinghua University}
  \city{Beijing}
  \country{China}
}
\email{dingjingtao186@qq.com}

\author{Zhicheng Liu}
\affiliation{%
  \institution{Tsinghua University}
  \city{Beijing}
  \country{China}
}
\email{zhichengliu@tsinghua.edu.cn}

\author{Yong Li}
\affiliation{%
  \institution{Tsinghua University}
  \city{Beijing}
  \country{China}
}
\email{liyong07@tsinghua.edu.cn}

\begin{abstract}
The Origin-Destination~(OD) networks provide an estimation of the flow of people from every region to others in the city, which is an important research topic in transportation, urban simulation, etc.
Given structural regional urban features, generating the OD network has become increasingly appealing to many researchers from diverse domains.
However, existing works are limited in independent generation of each OD pairs, i.e., flow of people from one region to another, overlooking the relations within the overall network.
In this paper, we instead propose to generate the OD network, and design a graph denoising diffusion method to learn the conditional joint probability distribution of the nodes and edges within the OD network given city characteristics at region level. 
To overcome the learning difficulty of the OD networks covering over thousands of regions, we decompose the original one-shot generative modeling of the diffusion model into two cascaded stages, corresponding to the generation of network topology and the weights of edges, respectively.
To further reproduce important network properties contained in the city-wide OD network, we design an elaborated graph denoising network structure including a node property augmentation module and a graph transformer backbone.
Empirical experiments on data collected in three large US cities have verified that our method can generate OD matrices for new cities with network statistics remarkably similar with the ground truth, further achieving superior outperformance over competitive baselines in terms of the generation realism. % The datasets and code are available online: https://anonymous.4open.science/r/KDD2023-20FB.
\end{abstract}


\maketitle

\input{Sec1_introduction}
\input{Sec2_relatedwork}
\input{Sec3_preliminaries}
\input{Sec4_method}

\input{Sec5_evaluation}

\section{Conclusion} \label{sec:Conclusion}
In this work, we propose to investigate the spatial distribution of the OD matrix, i.e. mobility flows between every two regions, of a city from the perspective of networks, and explore the feasibility of introducing the graph denoising diffusion method to model the joint distribution of all elements in the OD matrix. Specifically, we designed a cascaded graph denoising diffusion method (\textbf{DiffODGen}) to generate the city-wide OD matrix for the new city by first generating a topology structure and then mobility flows. By validating in two real-world data scenarios, the DiffODGen can effectively model the joint distribution of the city-wide OD matrix and generate the OD matrix with scaling network behaviours similar with real-world data to improve the precision. This demonstrates the necessity of modeling the joint distribution of all elements in the OD matrix and the feasibility of applying graph diffusion models to solve the problem of OD matrix generation.

% \clearpage

\bibliographystyle{ACM-Reference-Format}
\bibliography{bibliography}

% \appendix
% \input{Secend_appendix}

\end{document}

%% file: Sec1_introduction.tex
\vspace{-0.5cm}
\section{Introduction}
The ability to model population mobility for traffic control~\cite{arentze2004learning}, urban planning~\cite{vuchic2002urban} and resource scheduling~\cite{ruan2020dynamic,caminha2017impact} has critical impacts on everyday functioning and sustainable development of our cities. 
The Origin-Destination~(OD) matrix is a commonly used way of structuring mobility flow information in a city, which includes the number of travelers between every two regions. However, it costs a lot to gather the OD matrix of a new city through traditional travel surveys. This has led to studies that generate the OD matrices for new cities without any flow information based on city characteristics including geographic structure as well as the demographics and urban functions. Unlike OD matrix forecasting~\cite{wang2019origin,shi2020predicting}, this problem of city-wide OD matrix generation aims to generate the OD matrix for a city without any historical flow information.

Existing works for the OD matrix generation can be divided into two categories, i.e., physics-based methods~\cite{zipf1946p,simini2012universal,barbosa2018human} and machine learning-based methods~\cite{robinson2018machine,pourebrahim2019trip,liu2020learning,simini2021deep}. The physics-based methods draw an analogy between population and physical phenomena \cite{zipf1946p,simini2012universal} and model the flow of people between two regions through equations with several parameters. The machine learning-based methods follow the data-driven paradigm and utilize complex models with many parameters trained on a large amount of data to model the nonlinear dependencies between the mobility flow and features of the origin as well as the destination. Because of the complexity of population mobility, physics-based methods have relatively poor performance, while machine learning-based methods have recently achieved better results and wider applications~\cite{adnan2016simmobility,horl2021simulation,zeng2022causal,karimi2019sustainable} aided by the sophisticated model structure. 

Despite their capability of incorporating complex urban factors related to flow generation, existing works neglect the relations between elements in the city-wide OD matrix and instead generate each element, i.e., mobility flow, in an independent manner. In this regard, previous studies~\cite{yan2017universal,saberi2017complex} have investigated urban travel demand patterns in a network perspective and empirically observed the scaling behaviour, i.e., power laws, in distributions of node flux and link weight, etc. This important network property of the city-wide OD matrix~\cite{balcan2009multiscale,barthelemy2011spatial,yan2017universal} is vital for its generation realism and thus calls for a graph generative modeling solution that has been untouched in previous works.

Recently, deep generative models~\cite{kingma2013auto,goodfellow2020generative,ho2020denoising,song2019generative} are proposed to model complex joint distribution of data. Among them, the best-performing work is diffusion models~\cite{ho2020denoising}. Diffusion models approximate complex data distributions by gradually removing small noise from simple distributions~\cite{ho2020denoising}. The elaborate denoising process allows diffusion models to effectively fit complex distributions and generate high quality data. 
Therefore, we propose to solve the problem of city-wide OD matrix generation based on city characteristics via graph denoising diffusion methods.

However, it is hard to directly model joint distribution of the city-wide OD matrix via graph diffusion methods and generate the OD matrix for a new city. There are two challenges as follows. 

\begin{itemize}
    \item \textbf{How to overcome the difficulty of generating the city-wide OD matrix covering thousands of regions?} The city-wide OD matrix is equivalent to a mobility flow network with thousands of nodes, while existing deep generative graphic models mainly focus on the molecular structure~\cite{niu2020permutation,haefeli2022diffusion,jo2022score,vignac2022digress} and can only handle topologies with tens or hundreds of nodes. This means building continuous connections at million-level, which greatly increases the complexity of the joint distribution. 
    \item \textbf{How to achieve realistic reproduction of network properties contained in the mobility flow pattern in generation?} The mobility flow network/OD matrix exhibits sparsity and scaling behaviours. People prefer to travel a short distance, which causes no flow between most regions. Moreover, the spatial distribution of mobility in a city appears highly heterogeneous with scaling behavior on node and edge level~\cite{saberi2017complex,yan2017universal}. Maintaining these properties in generated city-wide OD matrices is a claim of realism, but very challenging.
\end{itemize}

Solving above two challenges calls for a decoupling of the topology and weights of the mobility flow network and locally graphic modeling from node and edge level in graph diffusion methods. We propose a cascaded graph denoising \textbf{\underline{diff}}usion method for soling the problem of city-wide \textbf{\underline{OD}} matrix \textbf{\underline{gen}}eration (\textbf{DiffODGen}). To generate the city-wide OD matrix and overcome the first challenge, we dismember the generation process into two stages, and construct a pipeline comprised by two diffusion models to deal with sub-tasks in each stage. In the first stage, our proposed method determines the existence of flow, i.e., topology structure of the mobility flow network with a topology diffusion model. Then, the flow volumes of region pairs that have flows will be generated via the flow diffusion model. For better integration of the two stages, we design a collaborative training to eliminate the cascading errors. For maintaining network properties of sparsity and scaling behaviors, we utilize the discrete denoising process~\cite{austin2021structured,vignac2022digress} in the topology diffusion model and apply a graph transformer-based network parameterization to model the city-wide OD matrix from network perspective respectively. Besides, we specially design node properties augmentation modules in both diffusion models to enhance the denoising networks with the capability of fully modeling the scaling behaviors on node level.

In summary, the contributions can be summarized as follows.
\begin{itemize}
\item  We propose to leverage the idea of graph generative modeling in learning the joint distribution of mobility flows for generating city-wide OD matrices in new cities. 
To the best of our knowledge, we are the first to resolve this important urban computing problem of OD matrix generation within the diffusion model paradigm.
\item  We design a cascaded graph denoising diffusion method and graph transformer network parameterization with a node properties augmentation module to generate city-wide OD matrix for new cities with the network properties retained.
\item Experiment results on two real-world dataset collected from two large US cities demonstrate the superiority of our DiffODGen over state-of-the-art baselines in terms of generating realistic city-wide OD matrices. Importantly, the generated OD matrix exhibits excellent network statistics similarities of sparsity and scaling behaviors in both node level and edge level with the real-world data.
\end{itemize}

%% file: Sec2_relatedwork.tex
\section{RELATED WORK} \label{sec:Relatedwork}
In this section, we will give a comprehensive review of the related works on OD Matrix generation and recent achievements of diffusion models in the field of graph learning.
\subsection{OD Flow Generation}
Research related to OD matrix generation has a very long history, with a recent boom brought about by the rise of machine learning algorithms. The methods utilized in these works are mainly in two categories. The first kind of methods are inspired by classic physical laws. In 1946, Zipf~\cite{zipf1946p} introduced Newton's physics law of Gravitation to model the mobility flow between two regions. The population of one region is considered as the mass while the flow between two regions is modeled as the universal gravitational force between objects. Simini et al.~\cite{simini2012universal} compare the movement of people within urban space to the processes of emission and absorption of radiation in solid physics. Due to the limitation of only modeling the complex population mobility based on simple physical models, these physical methods perform poorly.

With the advances in machine learning, a growing spectrum of problems are being tackled in the data-driven paradigm, and population mobility modeling is no exception. Robinson et al.~\cite{robinson2018machine} show that the tree-based machine learning methods perform much better than traditional physics-based methods, especially Gradient Based Regression Trees (GBRT). Pourebrahim et al.~\cite{pourebrahim2019trip} compare popular machine learning models and conclude that random forest performs best. Simini et al.~\cite{simini2021deep} introduce deep neural networks to enhance the gravity model with richer urban features, granting improved nonlinear modeling capabilities. Liu et al.~\cite{liu2020learning} bring the graph neural networks into the domain of OD flow prediction, using multitask learning as a training strategy to learn embedding with stronger representational power to improve the prediction. However, these models all only consider the features of one OD pair while ignoring complicated dependencies between mobility flows. These models cannot model the network statistic properties of the city-wide mobility flow networks~\cite{saberi2017complex}.

\subsection{Diffusion Models}
Diffusion models, which consist of a diffusion (noise-adding) process without any learning parameters and a reverse denoising process converting the sampled noise into data following complicated distribution with the help of denoising networks, are a kind of recent emerging generative models~\cite{ho2020denoising,song2019generative,nichol2021improved,song2020denoising}. Such models have outperformed other deep learning generative models such as Variational Auto-encoders (VAEs)~\cite{kingma2013auto}, Generative Adversarial Networks (GANs)~\cite{goodfellow2020generative} etc in many domains such as computer visions~\cite{ho2020denoising,saharia2022photorealistic}, audio generations~\cite{kong2020diffwave,tashiro2021csdi} and the graph generation~\cite{niu2020permutation,jo2022score,vignac2022digress}.

The most relevant works are those that apply diffusion models to graph generation task~\cite{niu2020permutation,jo2022score,vignac2022digress}. Niu et al.~\cite{niu2020permutation} are the first to utilize score-based diffusion models which adopt continuous Gaussian noise to construct the diffusion process to generate the adjacency matrix with the capability of permutation invariance. Further, Jo et al.~\cite{jo2022score} explore the possibility of jointly generating graphs with node and edge features employing diffusion models. Recent works~\cite{haefeli2022diffusion,vignac2022digress} have explored the construction of discrete denoising diffusion models with multinomial noise, enabling the generated graphs to retain sparsity and improve the generation quality. Our work explores the feasibility of applying graph diffusion models to generate the city-wide OD matrices (mobility flow networks), where not only the volumes of the mobility flow are generated by denoising continuous noise, but also the sparsity and scaling properties of OD matrices~\cite{saberi2017complex,yan2017universal} are ensured by graph denoising diffusion models.

%% file: Sec3_preliminaries.tex
\section{Preliminaries} \label{sec:pre}
In this section, we will list important notations and definitions. And a brief introduction to principles of diffusion models are presented.

\subsection{Definitions and Problem Formulation}

\begin{definition}
    \textbf{Regions.} The city is spatially partitioned into several non-overlapping regions denoted as $\mathcal{R}=\{ r_i | i=1,2,...,N \}$, where $N$ is the number of regions. The different regions are located in different parts of the city and perform different urban functions, which are reflected through the urban characteristics $\mathbf{X}_r$, such as demographics and points of interests (POIs).
\end{definition}
\begin{definition}
    \textbf{OD flow.} The OD flow denotes the directed mobility flow $\mathcal{F}_{r_{org},r_{dst}}$ between a specific region pair that departs from the origin $r_{org}$ and moves to the destination $r_{dst}$.
\end{definition}
\begin{definition}
    \textbf{OD Matrix.} The OD matrix $\mathbf{F} \in \mathbf{R}^{N \times N}$ includes the OD flows between every two regions in the city, where $F^{i,j}$ stands for the mobility flow from region $r_i$ to region $r_j$.
\end{definition}

\textsc{Problem 1.}
\textit{\textbf{OD Matrix Generation.} Given the regional urban characteristics of the city $ \{ \mathbf{X}_r | r \in \mathcal{R} \}  $, generate the whole picture of mobility flow of that city, i.e., the OD matrix $\mathbf{F}$.}

% \subsection{Network Perspective}
The OD matrix of a city can also be regarded as a directed weighted network $\mathcal{G}=\{\mathcal{V}, \mathcal{E}\}$, where the nodes $\mathcal{V}=\{v\}$ represent regions $\mathcal{R}=\{r\}$ and the directed edges $\mathcal{E}=\{e\}$ with weights $\{w_e\}$ represent mobility flows.

\begin{definition}
    \textbf{Mobility Flow Network.} The OD matrix $F$ in this paper has the same meaning as a mobility flow network $\mathcal{G}$, except that one is in terms of data organization and the other is from the network perspective. The element $F_{ij}$ is equivalent to $w_{e_{ij}}$, where $e_{ij}$ denotes the directed edge from node $i$ to node $j$.
\end{definition}
\begin{definition}
    \textbf{Adjacency Matrix.} A adjacency matrix $M$ is a 0-1 binary matrix, where $M_{ij} = 0$ means no edge from node $i$ to node $j$, i.e., no mobility flow from region $r_i$ to region $r_j$ and $M_{ij} = 1$ is the opposite. This indicates whether there are edges between nodes in the mobility flow network.
\end{definition}

\subsection{Diffusion Models}
The diffusion models, which consist of a diffusion process and a reverse denoising process, aim to learn the sophisticated data distribution $p(x^0)$ by removing noise from simple distributions using a Markov process~\cite{ho2020denoising}. Through the forward process $q(x^{t}|x^{t-1})$, the model produces a series of noisy data as hidden states $\{x^t|t=1,...,T\}$ by adding a small noise based on the original raw data $x^0$.
\begin{equation}\label{eq:diffprocess}
\begin{split}
\begin{aligned}
    & q(x^t|x^{t-1}) = \mathcal{N}(x^t; \sqrt{1-\beta_t} x^{t-1} , \beta_t \mathbf{I}),\\
    & q(x^1, ..., x^T|x^0) = \prod_{t=1}^T {q(x^t|q^{t-1})},
\end{aligned}
\end{split}
\end{equation}
where $T$ is the diffusion steps and $\{ \beta_t | t=1,...,T \}$ is variance schedule. When sampling data from learned distribution, we need a neural network $\theta$ to complete the reverse denoising process by predicting hidden states $p_{\theta}(x^{t-1}|x^t)$ without $x^0$ by neural networks until the last step of the denoising process has been performed:
\begin{equation}
    p_{\theta}(x^0,...,x^{T-1}) = \prod_{t=1}^{T} p_{\theta}(x^{t-1}|x^t),
\end{equation}
Different diffusion processes utilize different kinds of noises (Gaussian noise for continuous data~\cite{ho2020denoising} and multinomial noise for discrete categorical data~\cite{austin2021structured}), depending on the specific scenario.

%% file: Sec4_method.tex
\section{Method} \label{sec:methods}
In this section, we will detail the cascaded graph denoising diffusion method for city-wide OD matrix generation in new cities (DiffODGen). First, we will introduce the framework of the method and dive into concrete information about each part of it.

\subsection{Cascaded Graph Denoising Diffusion}

\subsubsection{\textbf{Overall Framework.}}

\begin{figure*}[t]
    \centering
    \includegraphics[width=0.80\textwidth]{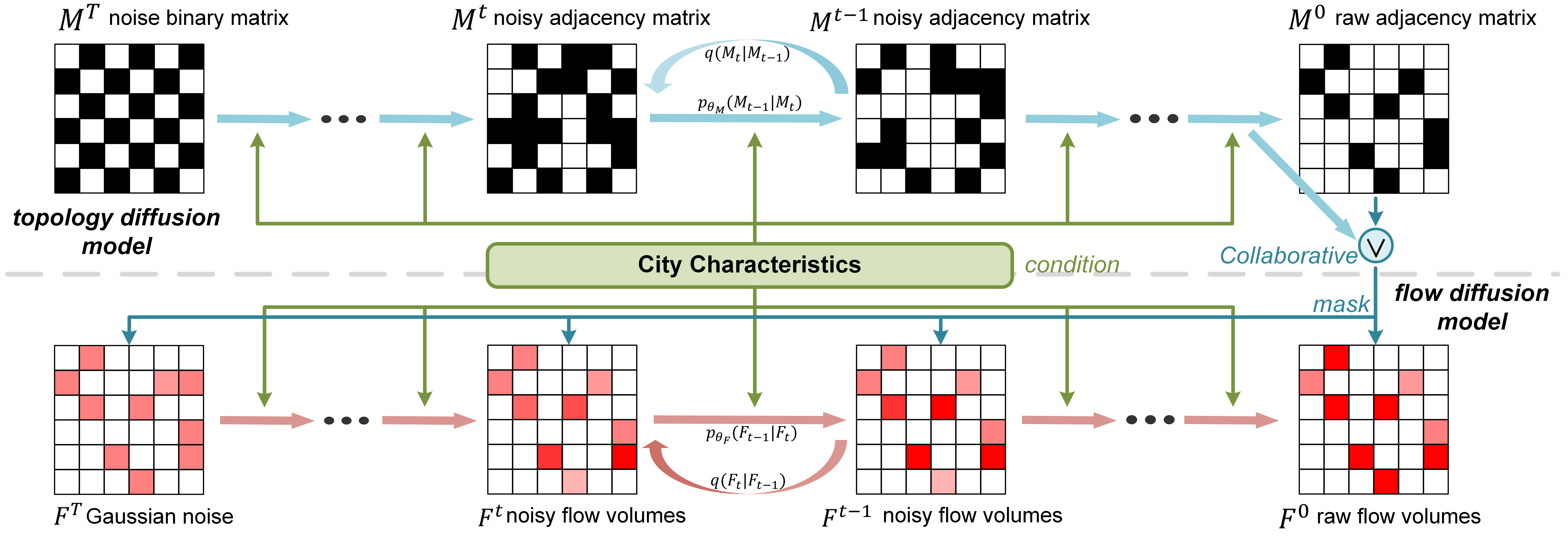}
    \vspace{-0.4cm}
    \caption{The architecture of our cascaded graph denoising diffusion method for the city-wide OD matrix generation.}
    \vspace{-0.5cm}
    \label{fig:framework}
\end{figure*}

As shown in Figure~\ref{fig:framework}, our method includes two stages, which can be treated as separating the city-wide OD matrix generation task into two steps. In the first stage, as we can see in the upper part of Figure~\ref{fig:framework}, the method determines whether there is a mobility flow between regions $P_{\theta_{M}}(M|X)$, i.e., topology structure, with the topology diffusion model. Then, the region pairs without flow will be directly set to zero flow and discarded in the rest of the process. According to the sparsity of the OD matrix, this will greatly reduce the scale of city-wide OD matrix generation and thus solve the first challenge. In the second stage, we use a continuous denoising diffusion model to learn the joint probability distribution of OD flows between the remaining region pairs $P_{\theta_{F}}(F|X)$ for generation.

The diffusion models in both the two stages are in the form of classifier-free conditional diffusion models~\cite{ho2022classifier}, where pairwise regional attributes are element-wise assigned to the OD flows between each two regions in the OD matrix. This modeling can make the generated OD matrix not only globally conform to the overall properties, but also retain the advantages of the pairwise predictive model by fully considering the features of OD pairs so that OD flows are also locally reasonable.

\subsubsection{\textbf{Collaborative Training.}}
\label{sec:cotrain}

A noteworthy issue is that combining the two diffusion models directly has the problem of cascading errors, i.e., the errors generated in the topology generation stage are completely transferred to the flow generation stage, which leads to an amplification of the overall instability. To solve this issue, we adopt the strategy of \textit{\textbf{teacher-force learning}} to collaborate the training of networks $\theta_{M}$ and $\theta_{F}$. The training process is based on data from the source city, where the true value of the OD matrix is known. Not the adjacency matrix $\hat{M}$ generated is applied for training the network parameters in the flow diffusion model, but the \textit{union} $\widetilde{M}$ of generated $\hat{M}$ and the ground truth $M$ instead, which is computed as follow,
\begin{equation}\label{eq:teacher}
    \widetilde{M}_{ij} = \hat{M}_{ij} \vee M_{ij}{\;}{\;}{\;}for{\;}i=1,...,N{\;}and{\;}j=1,...,N{\;},
\end{equation}
where the $\vee$ denotes the logic operator of \textit{OR}. $\widetilde{M}$ then serves as a mask in the flow diffusion model, so that it focuses only on the elements in the OD matrix whose value are not zero. In this way, the flow diffusion model has the capability of tolerating errors arising in the topology diffusion model and remedy them by generating the missing zero flows during the topology generation, thus improving the overall robustness of the method.

\subsection{Topology Diffusion Model}
Next, we will give a introduction to the topology diffusion model, including discrete denoising process and network parameterization.

\subsubsection{\textbf{Discrete Denoising Process.}}
To avoid destroying the sparsity of the generated OD matrix, we utilize the reverse denoising process with discrete state spaces to generate the adjacency matrix in the topology diffusion model. We adopt the d3pm proposed by Austin et al~\cite{austin2021structured}, which works for graph generation task~\cite{vignac2022digress}, to approximate the data distribution $p_M(M)$ with $p_{\theta_{M}}(M^0)$. In our approach, data $M_{ij} \in \mathbb{R}^d$, which follows a binary distribution, is encoded with one-hot encoding, where $d$ equals two, i.e., the number of classes (zero/non-zero flow). For the convenience of representation, we briefly refer to the element in adjacency matrix $M_{ij}$ as $m$ in the following. The noise-addition in the diffusion process is implemented through the matrix product with a series of transition matrix $(\mathbf{Q}^1,...,\mathbf{Q}^T)$,
\begin{equation}
    q(m^t|m^{t-1})=m^{t-1}\mathbf{Q}^t,
\end{equation}
such that $\mathbf{Q}^t_{ij}$ denotes the probability of state transiting from class $i$ to class $j$. It is worth noting that the noisy data state $m^t$ at $t$ diffusion step can be directly obtained by closed-form~\cite{austin2021structured} calculations,
\begin{equation}
    q(m^t|m^0) = m^0 \bar{\mathbf{Q}}^t,
\end{equation}
where $\bar{\mathbf{Q}}^t = \mathbf{Q}^1\mathbf{Q}^2...\mathbf{Q}^t$. The transition matrix $\textbf{Q}^t$ applied in our method introduce the uniform noise perturbation, which is calculated as follows,
\begin{equation}
    \mathbf{Q}^t = \alpha ^t \mathbf{I} + \frac{(1-\alpha ^t) {\mathbf{1}_d} {\mathbf{1}_d}^\top}{d},
\end{equation}
where $\top$ means the matrix transposition and $\alpha$ controls the degree of adding noise. Given $m^0$, the reverse denoising process can also be computed in closed-form based on Bayes rule~\cite{austin2021structured},
\begin{equation}
    q(m^{t-1}|m^t,m^0) = \frac{ m^t {\mathbf{Q}^t}^\top \odot m^0 \bar{\mathbf{Q}}^{t-1} }{m^0 \bar{\mathbf{Q}} {m^t}^\top},
\end{equation}
where $\odot$ means the element-wise product. Since $m^0$ is unknown, we construct neural networks to fit the probability of $p_\theta(\hat{m}^0|M^t)$ given noisy topology matrix $M^t$ and realize the denoising process with predicted $\hat{m}^0$ when doing generation,
\begin{equation}
    p_{\theta_{M}}(M^{t-1}|M^t){\;}\propto{\;} \sum_{\hat{m}^0} q(m^{t-1}|m^t,\hat{m}^0)p_{\theta_M} (\hat{m}^0|M^t).
\end{equation}
For generating the adjacency matrix based on specific city characteristics, we then extend the above denoising process into a conditional denoising process to model conditional probability $p_{\theta_{M}}(M^0|\mathbf{X}_{r\in \mathcal{R}})$. The conditional denoising process is as follows,
{\setlength\abovedisplayskip{0.2cm}
\setlength\belowdisplayskip{0.1cm}
\begin{equation}\label{eq:pm}
    p_{\theta_{M}}(M^{t-1}|M^t){\;}\propto{\;} \sum_{\hat{m}^0} q(m^{t-1}|m^t,\hat{m}^0)p_{\theta_M} (\hat{m}^0|M^t, \mathbf{X}_r).
\end{equation}
}

In the procedure of generation, a pure noise $M^T \in \mathbb{R}^{N \times N \times d}$, which follows the uniform distribution, is first sampled. The data samples that follow the original data distribution are then recovered gradually from the $M^T$ by the iterative calculation of reverse denoising process described by Eq.~\ref{eq:pm}, where $\theta_M$ is the trained neural networks. It should be emphasized that the whole process of recovering data through latent states is performed on a discrete space, which can fully preserve the sparsity of the OD matrix~\cite{vignac2022digress}.

% \vspace{-2cm}
\subsubsection{\textbf{Backbone of Network Parameterization.}}
For better considering the adjacency matrix, i.e., topology structure and fully modeling the network properties~\cite{saberi2017complex}, the network parameterization of topology diffusion model is designed to build with the graph transformer proposed by Dwivedi et al~\cite{dwivedi2020generalization}. Moreover, we make two improvements to the network, adapting it to classifier-free element-wise conditional schema and integrating a node properties augmentation module to enhance the ability of modeling the network properties of mobility flows~\cite{saberi2017complex}.

\begin{figure*}[t]
    \centering
    \includegraphics[width=0.8\textwidth]{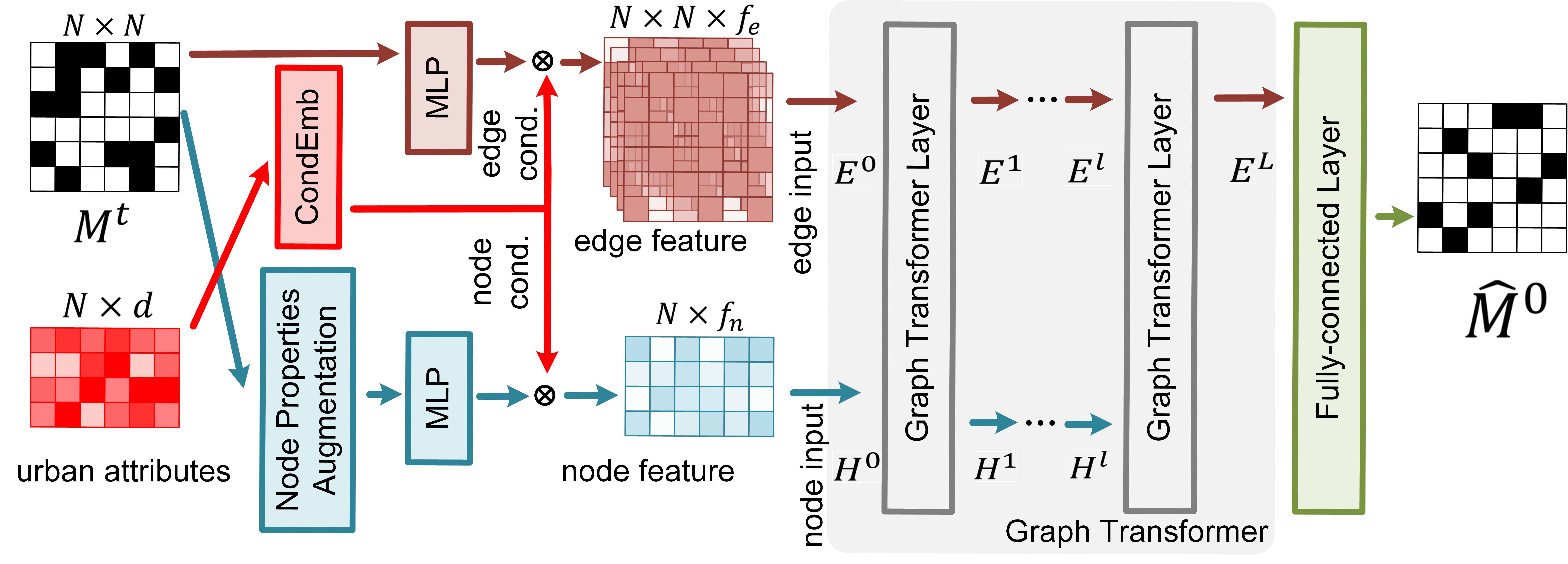}
    \vspace{-0.3cm}
    \caption{The architecture of network parameterization of the diffusion model.}
    \vspace{-0.3cm}
    \label{fig:netM}
\end{figure*}

The backbone of the graph transformer receives two inputs, node features and edge features, as shown in Figure~\ref{fig:netM}. The graph transformer consists of several layers. In each layer, all nodes form a position-insensitive sequence. The attention weights between each pair of nodes will be calculated through self-attention. It is worth noting that the calculation of attention weights used for information propagation between nodes also incorporates the influence of edge features. Each node updates its own features by weighted aggregation of neighbor information. Each edge updates itself by fusing its own features with the attention information between its end nodes. The calculation process of each layer in the graph transformer is as follows,
\begin{equation}
\begin{split}
\begin{aligned}
    h_i^{l+1} &= O_h^l \Vert_{k=1}^K ( \sum_{r_j \in \mathcal{N}_{r_i}} \alpha_{ij}^{k,l} V^{k,l} h_j^l ),\\
    e_{ij}^{l+1} &= O_e^l \Vert_{k=1}^K ( a_{ij}^{k,l} ), \\
    \alpha_{ij}^{k,l} &= softmax_j (a_{ij}^{k,l}), \\
    a_{ij}^{k,l} &= ( \frac{ Q^{k,l} h_i^l \cdot K^{k,l} h_j^l }{ \sqrt{d_k} } ) + W^{k,l} e_{ij}^l ,
\end{aligned}
\end{split}
\end{equation}
where $h_i^l$ means the hidden states of node $i$ after $l$ layers, $e_{ij}^l$ means the hidden states of the edge from node $i$ to node $j$ after $l$ layers, $O_h^l$, $O_e^l$, $Q^{k,l}$, $K^{k,l}$, $V^{k,l}$ and $W^{k,l}$ are the learnable parameters, $\Vert$ denotes the concatenation, $k=1$ to $K$ means the number of attention heads. The output of each layer will be input to the next layer. After the last layer, the output of node features $H^L$ is dropped and the output of edge features $E^L$ is converted into a tensor of the same shape as $M^t$ by a fully-connected layer to predict $\hat{M}^0$, as shown in Figure~\ref{fig:netM}.

\subsubsection{\textbf{Node Properties Augmentation for Topology.}}

Message-passing graph neural networks have a limitation of unable to capture some specific features~\cite{xu2018powerful,morris2019weisfeiler}, such as the weighted aggregation schema can hardly model the degree of nodes in a graph. However, the network properties of mobility flow are largely reflected in the region level, i.e., node level in mobility flow network~\cite{saberi2017complex}. Therefore, we augment the capability of our graph transformer with a \textit{node properties augmentation} to model the node level flow characteristics by statistically counting the properties of nodes into node features based on the adjacency matrix. The statistic properties of nodes include, but are not limited to, the degree of the node and the centrality, etc. With the help of node properties augmentation, the network can comprehensively model the noisy topology structure from both node and edge level in each diffusion step.

\subsubsection{\textbf{Classifier-free Element-wise Conditions.}}

As shown in Figure~\ref{fig:netM}, we make an improvement by introduce a \textit{CondEmb} module on the neural network backbone to make it classifier-free element-wise conditional denoising networks, so that diffusion models can determine whether there is a mobility flow based on its characteristics of origin and destination. First, urban attributes for each region $\{ \mathbf{X}_r | r \in \mathcal{R} \}$ are processed by MLPs to become regional condition embeddings, which have the same dimension as the node features that will be input to the graph transformer, as shown in Figure~\ref{fig:netM}. Then, regional condition embeddings are fused with node features and edge features respectively through cross attention $\otimes$ as the final input of the transformer, where the node features are directly fused with the embeddings region by region, while the edge features are fused with the concatenation of the embeddings of origin and destination as well as the distance between them. In this way, each row and column in the adjacency matrix, and each element has corresponding spatial urban features as its unique conditions that affect its generation in the denoising process.

\subsubsection{\textbf{Training of Topology Diffusion Model.}}

The network $\theta_M$ introduced above is trained by reducing the cross entropy between the prediction $\hat{M}_{ij}^0$ and true value $M_{ij}^0$ of each element to empower the denoising process represented by Eq.~\ref{eq:pm}. The overall cross entropy loss of the neural networks is calculated as follows,
\begin{equation}
    \mathcal{L}_M (\hat{M}^0,M^0) = \frac{1}{N^2} \sum_{1\leq i,j \leq N} \text{LCE}(p_{\theta_M}(\hat{M}^0_{ij} | M^t),M_{ij}),
\end{equation}
where LCE means cross entropy loss. The training algorithm of networks is shown as Algorithm~\ref{alg:AlgorithmT}. Once neural networks are trained, they can be employed to generate topology structure of mobility flow networks for better city-wide OD matrix generation.

\begin{algorithm}[t]
\caption{Training Algorithm of Topology Diffusion Model}
\label{alg:AlgorithmT}
\LinesNumbered
    \KwIn{\\
    \quad Cities characteristics of the training city:\\
    \begin{itemize}
        \item Spatial structure of the training city: $\mathcal{R}_{src}$
        \item Regional city characteristics: $\{ \mathbf{X}_{r} | r\in \mathcal{R}_{src} \}$
    \end{itemize}
    \quad Adjacency matrix of the training city: $\{ M_{ij} | r_i {\;} and {\;} r_j \in \mathcal{R}_{src} \}$
    }
    \KwOut{\\
    \quad Learned discrete denoising neural networks $\theta_M$.
    }
    
    \Repeat{Loss convergences.}{
        $M^0 = M$ \;
        Sample $t \sim \mathcal{U} (1,2,...,T)$ \;
        Sample $M^t \sim M^0 \bar{\mathbf{Q}}^t $ \;
        Predict $\hat{M}^0$ via $\theta_M$ given $M^t$ and $\{ \mathbf{X}_{r} | r\in \mathcal{R}_{src} \}$ \;
        $ loss \Longleftarrow \mathcal{L}_M (\hat{M}^0,M^0) $ \;
        optimizer.step($loss$) \;
    }
\end{algorithm}

\subsection{Flow Diffusion Model}

Given the adjacency matrix $\hat{M}^0$ generated by the topology diffusion model, the mask $\widetilde{M}$ that determines which region pairs the flows should be generated is calculated by Eq.~\ref{eq:teacher}. The mask $\widetilde{M}$ is then used in the flow diffusion model to determine which elements of the OD matrix need flow generation. It is worth noting that since the generated adjacency matrix is not completely accurate, there are still zero flows that will be generated by the flow diffusion model.

\subsubsection{\textbf{Continous Denoising Diffusion Models.}}

Different from topology generation, we adopt Gaussian noise to construct the diffusion process since OD flows are continuous values. Since DDPM~\cite{ho2020denoising} was proposed, diffusion models in continuous latent spaces have been explored to a significant extent~\cite{nichol2021improved}. In this part, the process of noise-addition diffusion of our diffusion models is performed on the image-like tensor, i.e., the OD matrix $F \in \mathcal{R}^{N \times N \times d}$, where $d$ equals 1. Differently, the reverse denoising process is performed only at the position, where the elements equal to 1, in the mask $\widetilde{M}$, as shown in Figure\ref{fig:framework}. Thanks to the sparsity of the OD matrix, although the OD matrix is large, our denoising process only focuses on the nonzero flow part, so we can effectively model it.

The diffusion process of the flow diffusion model is similar to the classical continuous diffusion models and has no parameters, which is shown in Eq.~\ref{eq:diffprocess} from right to left. So we will not cover it repeatedly, but rather introduce the denoising process in detail. The computation of each denoising step is shown as follows,
\begin{equation}
    p_{\theta_F} (F^{t-1}|F^t,\{\mathbf{X}_r\}) = \mathcal{N} (F^{t-1} ; \mu_{\theta_F} (X^t,t|\{\mathbf{X}_r\},(1-\bar{\alpha}^t)\mathbf{I}),
\end{equation}
where $\mathcal{N}$ means the Gaussian distribution, $\alpha$ is the noise schedule and $\bar{\alpha}^t = \alpha^1\alpha^2...\alpha^t$. 

\subsubsection{\textbf{Network Parameterization.}}

Similar to the topology diffusion model, we construct a conditional diffusion model by applying city characteristics as conditions to generate OD flows between regions. The network parameterization is also based on the graph transformer, except that we use the geo-contextual embedding learning methods proposed in GMEL~\cite{liu2020learning} as the CondEmb module shown in Figure~\ref{fig:netM} to enhance our condition features representational capabilities. To be specific, urban attributes $\{ \mathbf{X}_r | r \in \mathcal{R} \}$ for each region are processed by two GATs~\cite{velivckovic2017graph} to obtain the embeddings for each region as the origin and the destination, respectively. Then, region embeddings that are fused with the node features input into the graph transformer are composed of the origin embedding and the destination embedding of the region. The edge embeddings consist of the origin embedding of the origin and the destination embedding of the destination, as well as the distance between them. In addition, \textit{\textbf{node properties augmentation}} is also used \textit{\textbf{for flow generation}} to enhance the graph modeling capability from both node and edge level.

\begin{algorithm}[t]
\caption{Training Algorithm in Flow Diffusion Model}
\label{alg:AlgorithmF}
\LinesNumbered
    % input
    \KwIn{\\
    \quad Cities characteristics of the training city:\\
    \begin{itemize}
        \item Spatial structure of the training city: $\mathcal{R}_{src}$
        \item Regional city characteristics: $\{ \mathbf{X}_{r} | r\in \mathcal{R}_{src} \}$
        \item Mask $\widetilde{M}$ from the topology diffusion model
    \end{itemize}
    \quad OD matrix of the training city: $\{ F_{ij} | r_i {\;} and {\;} r_j \in \mathcal{R}_{src} \}$
    }
    \KwOut{\\
    \quad Learned noise prediction neural networks $\theta_F$.
    }

    \Repeat{Loss convergences.}{
        $F^0 = F$ \;
        Sample $t \sim \mathcal{U} (1,2,...,T)$ \;
        Sample $\mathbf{\epsilon} \sim \mathcal{N}(0, \mathbf{I}) $ \;
        $loss \Longleftarrow \left \| \widetilde{M} \cdot \left[ \mathbf{\epsilon} - \mathbf{\epsilon}_{\theta_F} (\sqrt{\bar{\alpha}^t} F^0 + \sqrt{1-\bar{\alpha}^t} \mathbf{\epsilon}, t, \{ \mathbf{X}_r \}) \right] \right \|^2 $ \;
        optimizer.step($loss$) \;
    }
\end{algorithm}

\subsubsection{\textbf{Training of Flow Diffusion Model.}}

Ho et al.~\cite{ho2020denoising} show that the denoising process can be trained by predicting the added noise in the diffusion process at the corresponding time step, i.e., optimizing the neural networks through the following objective,
\begin{equation}
    min_{\theta_F} \mathbb{E}_{F^0 \sim q(F^0), \mathbf{\epsilon} \sim \mathcal{N}(0, \mathbf{I}) } || \mathbf{\epsilon} - \mathbf{\epsilon_{\theta_F}} (F^t, t | \{ \mathbf{X}_r \}) ||_2^2,
\end{equation}
where $\mathbf{\epsilon}$ denotes the noise added during the step $t$ in diffusion process and $||\cdot||_2$ means the $L-2$ norm. The training algorithm is referred to Algorithm~\ref{alg:AlgorithmF}. The $\widetilde{M}$ is used in collaborative training schema to reduce the effect of cascade errors of combining the two stages, whcih has been introduced in Sec.~\ref{sec:cotrain}. Once the neural network is trained, it can be used to generate the OD flows for given topology of mobility flow network. First, a pure noise is sampled from the standard Gaussian noise $\mathcal{N}(0, \mathbf{I})$. The noise $\mathbf{\epsilon}_{\theta_F}$ is then sequentially predicted and removed from $F^t$ to obtain $F^{t-1}$. $F^0$ is finally generated via $T$ denoising step iteratively. It is worth emphasizing that elements with a mask of 0 in $\widetilde{M}$ are not involved in the training and generation process because of the mask $\widetilde{M}$.

%% file: Sec5_evaluation.tex
\section{Experiments} \label{sec:exp}

% In this section, we conduct extensive experiments to validate the superiority of our proposed two-phase denoising diffusion method on the OD matrix generation problem.

\subsection{Experimental Setup}

\subsubsection{\textbf{Data Description}}

\begin{table}[]
\setlength\tabcolsep{2pt}
 \centering
\caption{Basic statistics of the dataset of the cities~(``NZR$_\text{flow}$'' stands for the rate of nonzero flows).}\label{tab:cities}
 \vspace{-0.3cm}
\begin{tabular}{@{}c|ccccc@{}}
\toprule
City        & \multicolumn{1}{c}{\#regions} & \multicolumn{1}{c}{\#commuters} & \multicolumn{1}{c}{\#population} & \multicolumn{1}{c}{area(km$^2$)} & \multicolumn{1}{c}{NZR$_\text{flow}$} \\ \midrule
NYC           & 1,296                          & 1,694,884                         & 3,004,606                          & 451                           & 24.9\%                       \\ \midrule
Cook          & 1,319                          & 1,776,489                         & 1,974,181                          & 4,230                          & 25.2\%                       \\ \midrule
Seattle       & 721                           & 1,595,531                         & 3,495,493                          & 5,872                          & 41.9\%                       \\ \bottomrule
\end{tabular}
\end{table}

We choose three major US metropolises for our experiments. The cities are divided into regions at the census tract level. The basic statistic information on the cities is presented in Table~\ref{tab:cities}. The data for each city includes two parts, city characteristics and the OD matrix, which are introduced as follows.

\textbf{City Characteristics.} City characteristics depict the geographic structure of an entire city and the spatially heterogeneous distribution of city functions. In detail, each region carries attributes about demographics and urban functions, which are portrayed through American Community Survey Data collected by the U.S. Census Bureau and POIs distribution from OpenStreetMap~\cite{OpenStreetMap} respectively. Each region aggregates all the POIs therein to obtain a distribution over different categories. The distance between regions is determined in terms of the planar Euclidean distance between the centroids of the regions. Based on the above data processing, we extract features for each region, as a feature vector including demographic information by age, gender, income, etc. and the number of different categories of POIs, which are normally used for OD matrix generation as in previous works~\cite{robinson2018machine,pourebrahim2019trip,liu2020learning,simini2021deep}.

\textbf{Commuting OD Matrix.} The OD matrices are constructed based on the commuting flow collected in Longitudinal Employer-Household Dynamics Origin-Destination Employment Statistics (LODES) dataset in 2018. The commuting flows are aggregated into the regions. Each element in the OD matrices records the number of workers who are resident in one region and employed in another.

\subsubsection{\textbf{Baselines.}}
We compare our proposed DiffODGen with the following six baselines that can be categorized into three categories.

The first category contains two traditional methods.

% \begin{itemize}
\textbf{Gravity Model~(GM)}~\cite{barbosa2018human}. Motivated by Newton's law of Gravitation, GM assumes that the mobility flow is positively related to the populations of origin and destination, and negatively related to the distance between them.
    
\textbf{Random Forest~(RF)}~\cite{robinson2018machine,pourebrahim2019trip}. RF is a widely adopted tree-based machine learning model thanks to its robustness, achieving rather competitive performance in the task of OD flow generation.

The second category contains two state-of-the-art~(SOTA) deep-learning methods developed for OD flow generation.
    
\textbf{Deep Gravity Model~(DGM)}~\cite{simini2021deep}. DGM introduces deep neural networks into the modeling of gravity models to obtain flows by predicting the probability distribution to different regions when the outflow is given. We adapt the model to predict volumes of OD flow directly, making it to generate the OD matrix for new cities.

\textbf{Geo-contextual Multitask Embedding Layer~(GMEL)}~\cite{liu2020learning}. GMEL uses graph neural networks~(GNNs) to aggregate information from neighbors of each region~(node) to capture its spatial features in a city~(graph), which helps learn better-quality regional embeddings to achieve better prediction accuracy.

Finally, we additionally compare two deep generative models to check the validity  of our design in terms of generating a realistic mobility flow network, i.e., the OD matrix.

\textbf{NetGAN}~\cite{bojchevski2018netgan}. NetGAN  is a GAN-based model that mimics real networks by generating random walking sequences with the same distribution as those sampled from the real networks. We adapt it to generate directed weighted networks, i.e., OD matrices.
    
\textbf{Denoising Diffusion Probabilistic Model~(DDPM)}~\cite{ho2020denoising}. We also adopt the naive diffusion model that directly learns to generate the OD matrix as a baseline. In this baseline, no distinguishing treatment is assigned to zero elements and non-zero elements in the OD matrix.
% \end{itemize}

\begin{table*}[t]
 \setlength\tabcolsep{2pt}
 \centering
 \caption{OD matrix generation results on test cities. The best results are in bold and the next-to-the-best results are underlined.}
 \vspace{-0.3cm}
 \label{tab:performace}
\begin{tabular}{@{}c|ccc|ccc|ccc|ccc@{}}
\toprule
       Test City         & \multicolumn{6}{c|}{Cook County~($|\mathcal{R}|$=1.3$k$) $\longrightarrow$ New York City~($|\mathcal{R}|$=1.3$k$)}                                                                                                                                                                                                                                                                                              & \multicolumn{6}{c}{Cook County~($|\mathcal{R}|$=1.3$k$) $\longrightarrow$ Seattle~($|\mathcal{R}|$=0.72$k$)}                                                                                                                                                                                                                                                                                          \\ \midrule 
                
      \multicolumn{1}{c|}{\multirow{2}{*}{Model}}          & \multicolumn{3}{c|}{Flow Value Error}                                                     & \multicolumn{3}{c|}{Network Statistics Similarity}                                                                                                                                                                                        & \multicolumn{3}{c|}{Flow Value Error}                                                     & \multicolumn{3}{c}{Network Statistics Similarity}                                                                                                                                                                                        \\  
     \multicolumn{1}{c|}{}           & \multicolumn{1}{c}{RMSE$\downarrow$} & \multicolumn{1}{c}{NRMSE$\downarrow$} & \multicolumn{1}{c|}{CPC$\uparrow$}            & \multicolumn{1}{c}{$\text{JSD}_\text{inflow}$$\downarrow$} & \multicolumn{1}{c}{$\text{JSD}_\text{outflow}$$\downarrow$} & \multicolumn{1}{c|}{$\text{JSD}_\text{ODflow}$$\downarrow$} & \multicolumn{1}{c}{RMSE$\downarrow$} & \multicolumn{1}{c}{NRMSE$\downarrow$} & \multicolumn{1}{c|}{CPC$\uparrow$}            & \multicolumn{1}{c}{$\text{JSD}_\text{inflow}$$\downarrow$} & \multicolumn{1}{c}{$\text{JSD}_\text{outflow}$$\downarrow$} & \multicolumn{1}{c}{$\text{JSD}_\text{ODflow}$$\downarrow$} \\ \midrule
GM   & 4.806                    & 1.066                     & \multicolumn{1}{c|}{0.196}          & 0.684                                                                    & 0.399                                                                     & 0.407                                                                     & 12.709                   & 1.021                     & \multicolumn{1}{c|}{0.217}          & 0.592                                                                    & 0.370                                                                     & 0.482                                                                    \\
RF   & 5.717                    & 1.268                     & \multicolumn{1}{c|}{$\underline{0.361}$}          & $\underline{0.296}$                                                                   & $\underline{0.188}$                                                                     & 0.054                                                                     & 22.886                   & 1.839                     & \multicolumn{1}{c|}{$\underline{0.390}$}          & 0.068                                                                    & 0.154                                                                     & \textbf{0.028}                                                           \\ \midrule
DGM    & 4.453                    & 0.987                     & \multicolumn{1}{c|}{0.213}          & 0.511                                                                    & 0.598                                                                     & 0.047                                                                     & $\underline{11.738}$                   & $\underline{0.943}$                     & \multicolumn{1}{c|}{0.368}          & $\underline{0.040}$                                                                    & 0.366                                                                     & 0.076                                                                    \\
GMEL             & $\underline{4.427}$                    & $\underline{0.981}$                     & \multicolumn{1}{c|}{0.271}          & 0.450                                                                    & 0.656                                                                     & $\underline{0.040}$                                                                     & 12.415                   & 0.997                     & \multicolumn{1}{c|}{0.278}          & 0.416                                                                    & 0.333                                                                     & 0.514                                                                    \\ \midrule
NetGAN          & 4.546                    & 1.007                     & \multicolumn{1}{c|}{0.290}          & 0.478                                                                    & 0.396                                                                     & $\underline{0.040}$                                                                     & 12.617                   & 1.007                     & \multicolumn{1}{c|}{0.297}          & 0.396                                                                    & 0.315                                                                     & 0.426                                                                    \\
DDPM & 4.939                    & 1.094                     & \multicolumn{1}{c|}{0.290}          & 0.594                                                                    & 0.355                                                                     & 0.063                                                                     & 13.001                   & 1.044                     & \multicolumn{1}{c|}{0.379}          & 0.440                                                                    & $\underline{0.085}$                                                                     & 0.092                                                                    \\ \midrule
\multirow{2}{*}{DiffODGen}            & \multicolumn{1}{c}{\multirow{2}{*}{\begin{tabular}[c]{@{}c@{}}\textbf{4.288}\\ \footnotesize{~(+3.14\%)}\end{tabular}}}            & \multicolumn{1}{c}{\multirow{2}{*}{\begin{tabular}[c]{@{}c@{}}\textbf{0.950}\\ \footnotesize{~(+3.26\%)}\end{tabular}}}            & \multicolumn{1}{c|}{\multirow{2}{*}{\begin{tabular}[c]{@{}c@{}}\textbf{0.363}\\ \footnotesize{~(+0.60\%)}\end{tabular}}} & \multirow{2}{*}{\textbf{0.051}}                                                           & \multirow{2}{*}{\textbf{0.064}}                                                            & \multirow{2}{*}{\textbf{0.020}}                                                            & \multicolumn{1}{c}{\multirow{2}{*}{\begin{tabular}[c]{@{}c@{}}\textbf{10.109}\\ \footnotesize{~(+13.9\%)}\end{tabular}}}           & \multicolumn{1}{c}{\multirow{2}{*}{\begin{tabular}[c]{@{}c@{}}\textbf{0.812}\\ \footnotesize{~(+13.9\%)}\end{tabular}}}             & \multicolumn{1}{c|}{\multirow{2}{*}{\begin{tabular}[c]{@{}c@{}}\textbf{0.433}\\ \footnotesize{~(+11.0\%)}\end{tabular}}} & \multirow{2}{*}{\textbf{0.023}}                                                           & \multirow{2}{*}{\textbf{0.029} }                                                           & \multirow{2}{*}{$\underline{0.062}$}                                                                    \\
                    & \multicolumn{1}{c}{}                                                               & \multicolumn{1}{c}{}                   &     \multicolumn{1}{c|}{} & & &  & \multicolumn{1}{c}{}                                                               & \multicolumn{1}{c}{}                   &     \multicolumn{1}{c|}{} & & &              \\
\bottomrule
\end{tabular}
\end{table*}

\subsubsection{\textbf{Evaluation Metrics.}}
We use two kinds of evaluation metrics to verify the generation realism, including three metrics related to \textit{flow value error} and three metrics related to \textit{network statistics similarity}. Specifically, the three error-based metrics are Root Mean Square Error ({RMSE}), Normalized Root Mean Square Error ({NRMSE}) and the commonly adopted Common Part of Commuting ({CPC}).
\begin{equation}
    RMSE{\;}={\;}\sqrt{{\frac{1}{|\textbf{F}|}}{\sum\nolimits_{r_i,r_j\in{\mathcal{R}}}{||}{{{\textbf{F}}_{r_i,r_j}}-{\hat{\textbf{F}}_{r_i,r_j}}{{||}_2^2}}}},
\end{equation}
% \begin{equation}
%     NRMSE{\;}={\;}{ \frac{RMSE}{ \sqrt{ \frac{ \sum_{r\in{\mathcal{R}}}^{r_i,r_j} || F_{ij} - \bar{F}_{ij} ||_2^2 }{N^2} } } },
% \end{equation}
\begin{equation}
    NRMSE{\;}={\;}{ RMSE / \sqrt{ \frac{1}{N^2} \sum\nolimits_{r_i,r_j\in{\mathcal{R}}} || F_{ij} - \bar{F}_{ij} ||_2^2 } } ,
\end{equation}
% \begin{equation}
%     CPC{\;}={\;}{ \frac{2 \sum_{r\in{\mathcal{R}}}^{r_i,r_j} min(\mathbf{F}_{r_i,r_j}, \hat{\mathbf{F}}_{r_i,r_j})} {\sum_{r\in\mathcal{R}}^{r_i,r_j}{\mathbf{F}_{r_i,r_j}}+ \sum_{r\in\mathcal{R}}^{r_i,r_j} {\hat{\mathbf{F}}_{ij}} } },
% \end{equation}
\begin{equation}
    CPC={ 2\!\! \sum_{r_i,r_j\in{\mathcal{R}}}    \min(\mathbf{F}_{r_i,r_j}, \hat{\mathbf{F}}_{r_i,r_j}) / ({\sum_{r_i,r_j\in{\mathcal{R}}}{\!\mathbf{F}_{r_i,r_j}}+ \sum_{r_i,r_j\in{\mathcal{R}}}{\!\hat{\mathbf{F}}_{ij}} }) },
\end{equation}
where the $\bar{\mathbf{F}}$ denotes the mean. 
To calculate the statistical similarity between generated OD matrices and real OD matrices, we adopt Jensen-Shannon Divergence ({JSD}) and use it to measure the distance between distributions of generated data and real data, with respect to three typical network statistics, i.e., inflow, outflow and OD flow. The computation of {JSD} is shown as follow,

\begin{equation}
    JSD{\;}={\;} {{\textbf{KL}(\textbf{P}_{\mathbf{F}}||\textbf{P}_{\hat{\mathbf{F}}}) +  \textbf{KL}(\textbf{P}_{\hat{\mathbf{F}}}||\textbf{P}_{\mathbf{F}})}/{2}},
\end{equation}
where $\textbf{KL}$ means Kullback–Leibler divergence and $\mathbf{P}$ denotes the empirical probability distribution. The inflow and outflow are calculated by summing all flows into and out of regions, respectively.

\subsubsection{\textbf{Parameter Settings.}}

 The number of layers of graph transformer in the topology and flow generation phases are 2 and 3 respectively, while the numbers of channels are both 64. The diffusion steps of two diffusion models in our method are 1000, and both use a cosine noise scheduler proposed by Nichol et al~\cite{nichol2021improved}. The denoising networks in two diffusion models are trained with Adam optimizer~\cite{kingma2014adam} with a learning rate of 3e-4. 

The gravity model follows the way in~\cite{barbosa2018human} with 4 fitting parameters. The $n\_estimators$ of random forest are 100. The number of layers for DGM is 10. The number of layers in GNN-based models is 3 and the number of channels is 64. The DDPM baseline follow the parameter settings with the flow diffusion model in our method.

\subsection{Overall Performance}

\textbf{Performance comparison.}
In Table~\ref{tab:performace} we compare the performance of DiffODGen with baseline methods by training models on data collected in \textit{Cook County}~(in Chicago) and reporting generation results in two new cities, i.e., \textit{New York City} and \textit{Seattle}. From the results, we have the following findings:

\begin{itemize}[itemsep=2pt,topsep=0pt,parsep=0pt,leftmargin=*]
    \item \textbf{Our proposed cascaded graph denoising diffusion  method, i.e., DiffODGen, steadily achieves the best performance.}
    The OD matrix generated by DiffODGen achieves the best realism in terms of both flow value error and network statistics similarity. Specifically, compared with the best baseline, the flow value error~(RMSE/NRMSE) is reduced by over 3\% and 13\% in \textit{New York City} and \textit{Seattle}, respectively. Moreover, the statistical distribution similarity between the generated OD matrix and real one is rather high, indicated by a near-zero value of JSD.
    \item \textbf{Current deep-learning-based OD flow generation methods perform poorly in terms of network statistics similarity.} Aided by the capability of modeling nonlinear relationships between city characteristics and flows, deep-learning-based methods, i.e., DGM and GMEL, achieve next-to-the-best accuracy with respect to flow value error. However, the results of the three JSD metrics indicate that they cannot generate a realistic OD matrix regarding its network statistics, due to their independent modeling of each OD matrix element.
    \item \textbf{Deep generative methods cannot compete with traditional methods without considering the unique characteristics of the OD matrix.} By comparing NetGAN and DDPM, i.e., two generative models widely used in other domains, with RF, we observe that the former results have much larger JSD values, suggesting a less similar network generated by NetGAN and DDPM. For DDPM in particular, this naive diffusion model is hardly comparable to the rather simple GM due to the lack of the cascaded diffusion designed for handling city-wide OD matrix and the network property augmentation leveraged along the denoising process.
    \item \textbf{Despite the large deviation between the train and the test city, DiffODGen shows stable performance in terms of generation realism. } From Table~\ref{tab:cities} we can observe that the deviation between \textit{Seattle} and \textit{Cook} is larger than that between \textit{New York City}. Nevertheless, DiffODGen exhibits much significant advantage when tested in \textit{Seattle}, which is indeed more difficult. This demonstrates its capability of capturing the underlying mechanism behind the observed city-wide OD matrix, owing to its generative modeling from a network perspective.

\end{itemize}

\begin{figure}[t]
\centering
\subfigure[Inflow dist., New York City~($R^2$=0.89)]{
    \label{nycin}
    \includegraphics[width=0.22\textwidth]{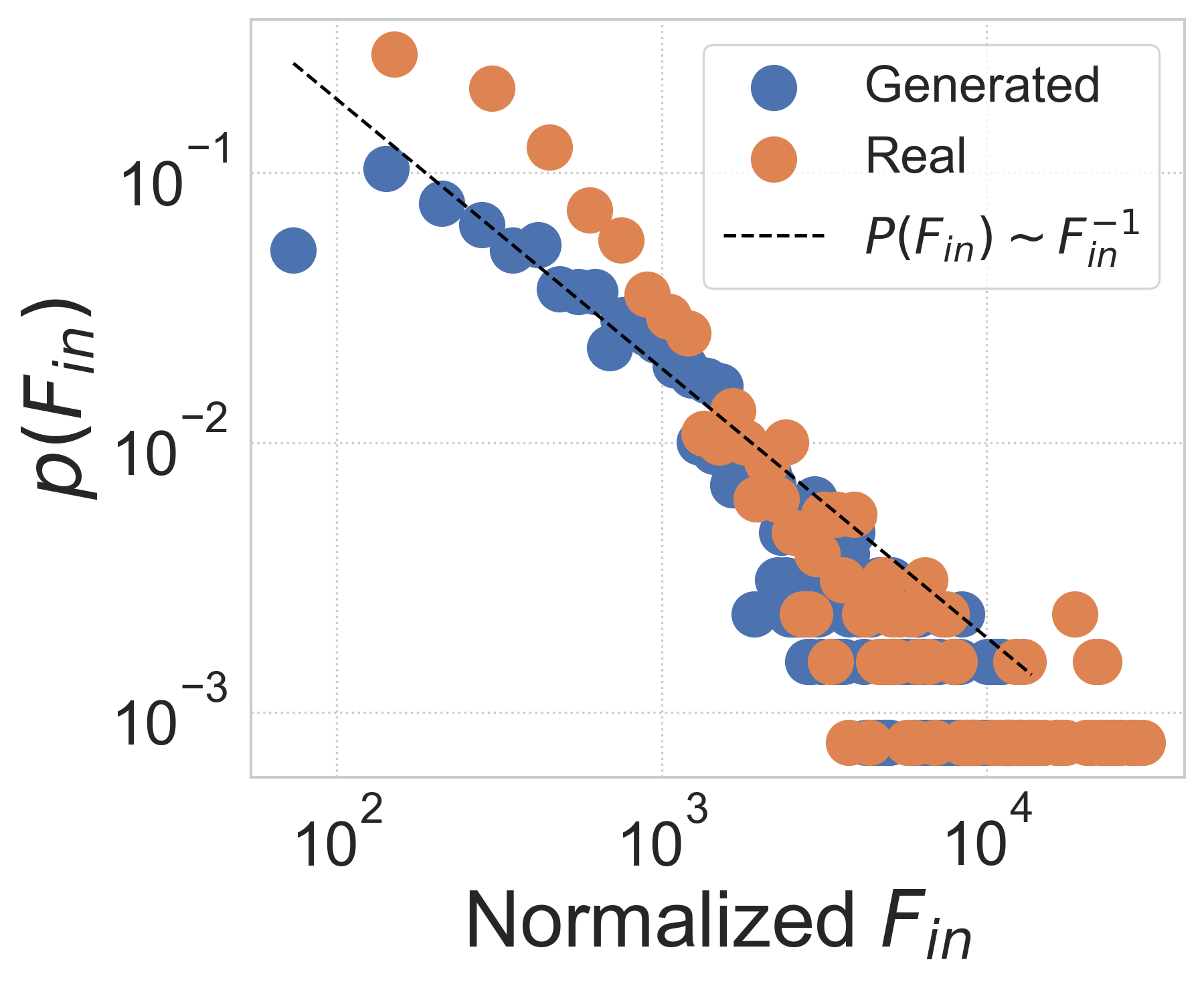}
}
\subfigure[OD flow dist., NYC~($R^2$=0.98)]{
    \label{nycout}
    \includegraphics[width=0.22\textwidth]{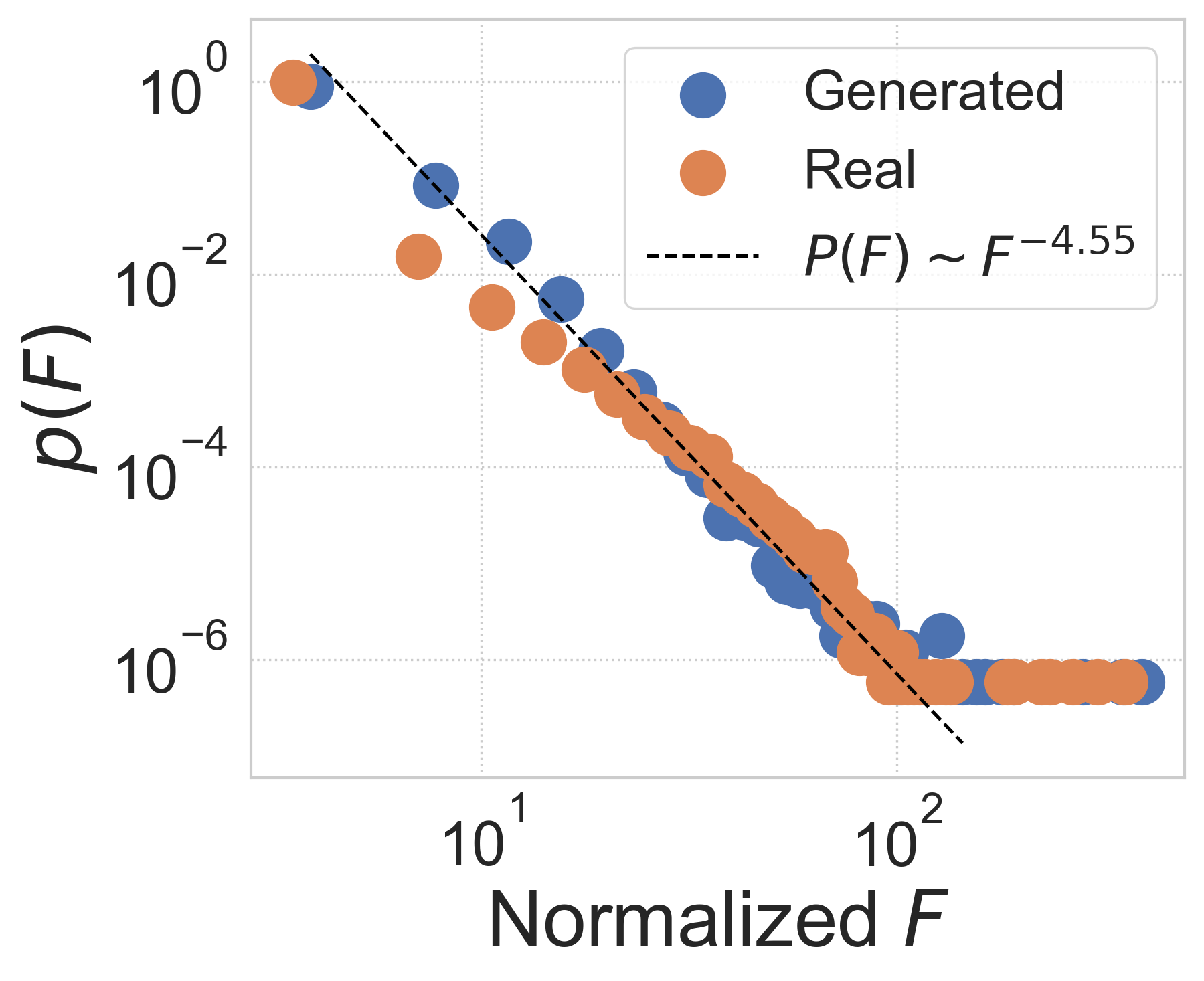}
}
\subfigure[Inflow dist., Seattle~($R^2$=0.86)]{
    \label{seain}
    \includegraphics[width=0.22\textwidth]{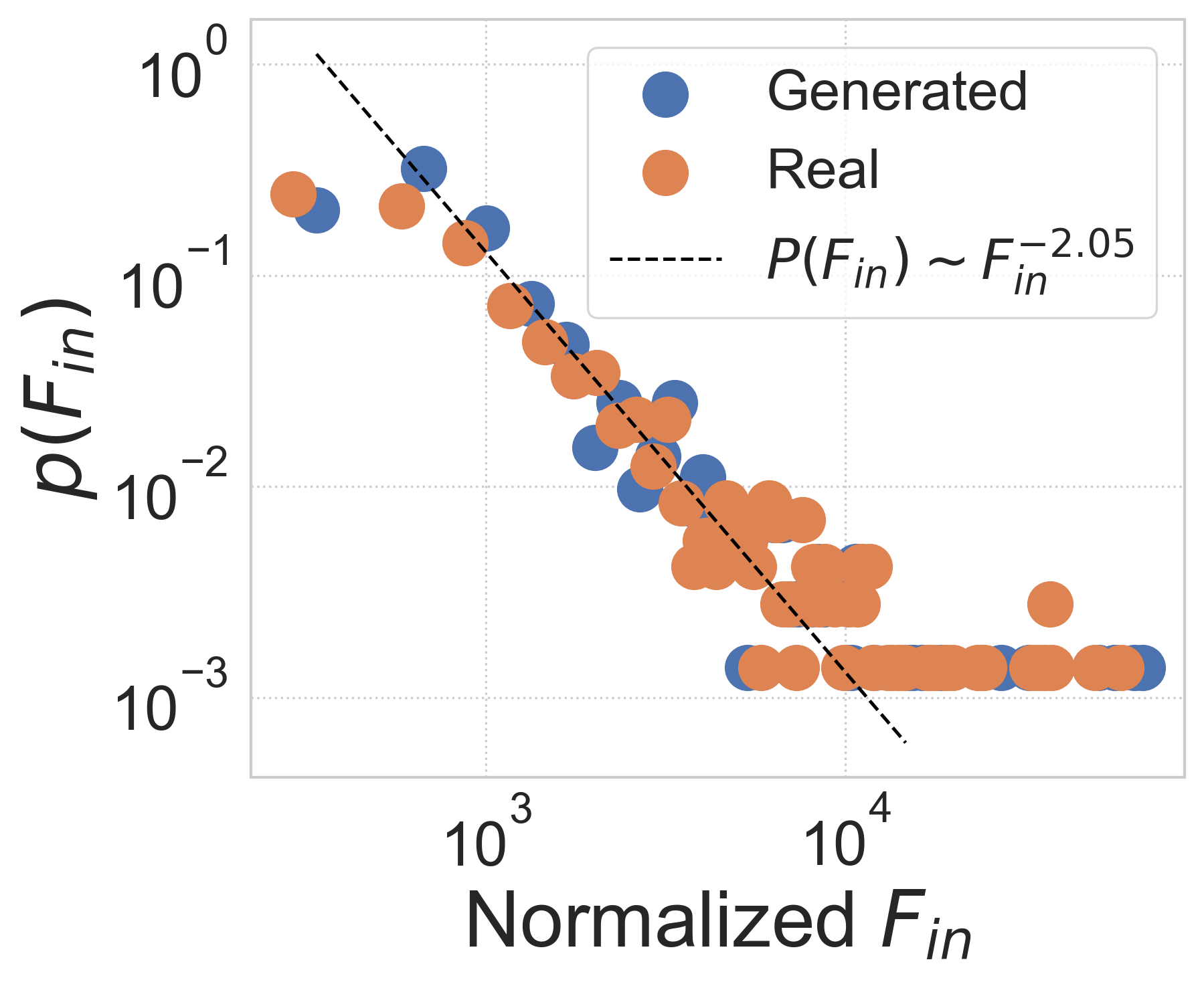}
}
\subfigure[OD flow dist., Seattle~($R^2$=0.97)]{
    \label{seaout}
    \includegraphics[width=0.22\textwidth]{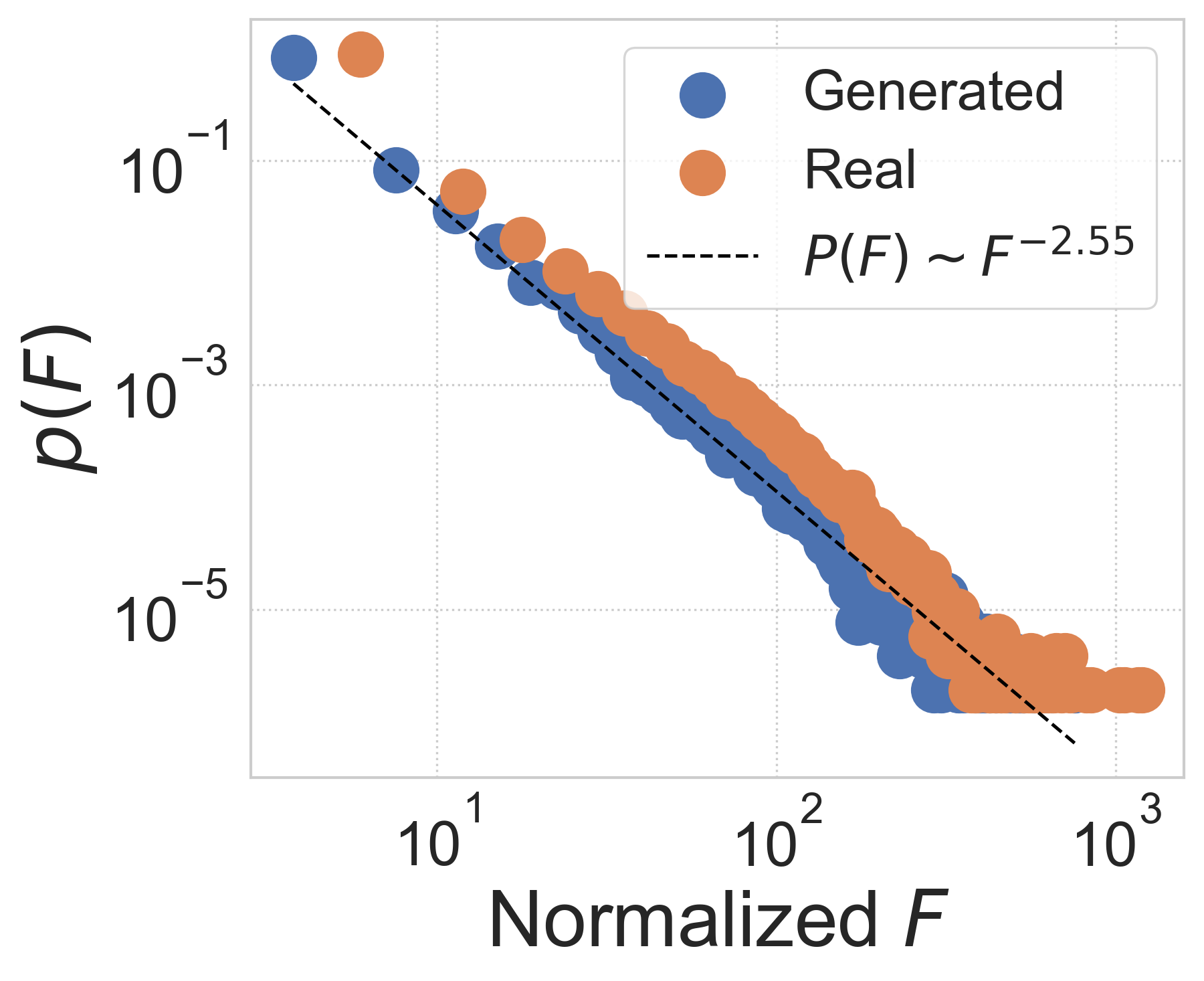}
}
\vspace{-0.4cm}
\caption{Distribution of two typical network statistics, i.e., inflow and OD flow, for both generated and real OD matrices of New York City and Seattle~(The dotted line indicates a pow-law fitting, with $R^2$ provided).}
\vspace{-0.5cm}
\label{scaling}
\end{figure}

\noindent\textbf{Network statistics analysis.}
We further investigate the realism of generated city-wide OD matrix from network perspective. Previous works on mobility flow networks have empirically observed scaling behavior in distributions, i.e., a pow-law-like distribution $P(x)\sim x^{-\alpha}$, of node-level inflow~\cite{saberi2017complex} and edge-level OD flow~\cite{yan2017universal}. Correspondingly, in Figure~\ref{scaling} we calculate probability distributions of these two statistics in both generated and real data, finding that the two distributions of inflow and OD flow show a remarkably good match to its power-law fitting, indicated by high values of $R^2$$\sim$0.86-0.98. Note that we normalize two metrics by their means for better visualization.
Moreover, as a heavy-tailed distribution like the power law indicates population heterogeneity that is generally hard to capture, the results demonstrate that DiffODGen successfully reproduces the network property contained in the mobility flow pattern by generating realistic city-wide OD matrices.

\subsection{{Topology Generation Performance}}

\begin{table}[t]
\setlength\tabcolsep{2pt}
 \centering
\caption{Performance comparison of topology generation results in two test cities~(NZR$_\text{flow}$ denotes the relative change ratio with respect to the original rate of nonzero flows in two cities).}\label{tab:topology1}
\vspace{-0.3cm}
% \resizebox{14cm}{!}{
\begin{tabular}{@{}c|ccc|ccc@{}}
\toprule
       Test City         & \multicolumn{3}{c|}{New York City~($|\mathcal{R}|$=1.3$k$)} & \multicolumn{3}{c}{Seattle~($|\mathcal{R}|$=0.72$k$)}      \\ \midrule           
      \multicolumn{1}{c|}{Model}          & \multicolumn{1}{c}{$\text{CPC}_\text{0/1}$} & \multicolumn{1}{c}{NZR$_\text{flow}$} & \multicolumn{1}{c|}{$\text{JSD}_\text{degree}$} & \multicolumn{1}{c}{$\text{CPC}_\text{0/1}$} & \multicolumn{1}{c}{NZR$_\text{flow}$} & \multicolumn{1}{c}{$\text{JSD}_\text{degree}$}  \\ \midrule
GM         & 0.356               & +253.4\%              & \multicolumn{1}{c|}{0.862}               & 0.585               & +139.0\%              & \multicolumn{1}{c}{0.815} \\
RF         & $\underline{0.476}$ & +103.6\%              & \multicolumn{1}{c|}{$\underline{0.486}$} & \textbf{0.647}      & $\underline{+43.4\%}$ & \multicolumn{1}{c}{0.064} \\
DGM        & 0.334               & $\underline{-72.3\%}$ & \multicolumn{1}{c|}{0.645}               & 0.618               & +52.4\%               & \multicolumn{1}{c}{$\underline{0.055}$} \\
GMEL        & 0.395               & +300.0\%              & \multicolumn{1}{c|}{0.842}               & 0.591               & +143.9\%              & \multicolumn{1}{c}{0.820} \\
NetGAN     & 0.411               & +281.1\%              & \multicolumn{1}{c|}{0.856}               & 0.597               & +126.1\%              & \multicolumn{1}{c}{0.792} \\ \midrule
DiffODGen  & \textbf{0.501}      & \textbf{+32.5\%}      & \multicolumn{1}{c|}{\textbf{0.135}}      & $\underline{0.630}$ & \textbf{-37.7\%}      & \multicolumn{1}{c}{\textbf{0.040}} \\ \bottomrule
\end{tabular}
% }
\vspace{-0.6cm}
\end{table}

To further demonstrate the necessity of leveraging graph generative modeling technique for city-wide OD matrix generation, we investigate the network topology generation performance of DiffODGen by comparing its first-stage results with other baseline methods in both \textit{New York City} and \textit{Seattle}.

\noindent\textbf{Performance comparison.}
We report the performance comparison results with respect to network topology similarity between generated and real OD matrices~(here we use the binary version, i.e., adjacency matrix) in Table~\ref{tab:topology1}. We choose the binary version of CPC~($\text{CPC}_\text{0/1}$), the rate of nonzero flows~(NZR$_\text{flow}$) and degree distribution similarity~($\text{JSD}_\text{degree}$).
\textbf{First}, results in \textit{New York City} demonstrate the superiority of DiffODGen in reconstructing the most realistic network topology given city characteristics in a new city. The generated mobility flow network achieves the highest $\text{CPC}_\text{0/1}$ indicating a good similarity to the real one, and they have close network statistics in terms of sparsity~(the smallest gap between NZR$_\text{flow}$) and degree distribution~(the lowest $\text{JSD}_\text{degree}$). Comparatively, we observe that other baselines cannot reproduce the realistic network topology. They tend to generate rather too sparse~(DGM) or too dense~(GMEL) networks.
\textbf{Second}, DiffODGen still generates the most realistic network topology in \textit{Seattle} when evaluating the overall three metrics. Note that RF performs slightly better in $\text{CPC}_\text{0/1}$, which may be attributed to the smaller scale~($|\mathcal{R}|$=0.72$k$) of \textit{Seattle} that can be easier for traditional modeling methods.

\begin{table}[t]
\renewcommand{\arraystretch}{0.8}
\setlength\tabcolsep{2pt}
 \centering
\caption{Further study of the network topology in generated city-wide OD matrices~(adjacency matrices).}\label{tab:topology2}
 \vspace{-0.3cm}
\begin{tabular}{@{}c|ccc|ccc@{}}
\toprule
       Test City         & \multicolumn{3}{c|}{New York City~($|\mathcal{R}|$=1.3$k$)} & \multicolumn{3}{c}{Seattle~($|\mathcal{R}|$=0.72$k$)}      \\ \midrule
       \multicolumn{1}{c|}{Model}          & \multicolumn{1}{c}{Accuracy} & \multicolumn{1}{c}{$\text{FN}_{\text{1}\rightarrow\text{0}}$} & \multicolumn{1}{c|}{$\text{FP}_{\text{0}\rightarrow\text{1}}$} & \multicolumn{1}{c}{Accuracy} & \multicolumn{1}{c}{$\text{FN}_{\text{1}\rightarrow\text{0}}$} & \multicolumn{1}{c}{$\text{FP}_{\text{0}\rightarrow\text{1}}$}  \\ \midrule
      % \multicolumn{1}{c|}{Model}          & \multicolumn{1}{c}{Accuracy} & \multicolumn{1}{c}{FN~($1 \rightarrow 0$) & \multicolumn{1}{c|}{FP~($0 \rightarrow 1$} & \multicolumn{1}{c}{Accuracy} & \multicolumn{1}{c}{FN~($1 \rightarrow 0$) & \multicolumn{1}{c|}{FP~($0 \rightarrow 1$}  \\ \midrule
GM         & 0.276                & 0.183                & \multicolumn{1}{c|}{0.900}               & 0.419               & $\underline{0.022}$ & \multicolumn{1}{c}{0.984}               \\
RF         & 0.605                & 0.271                & \multicolumn{1}{c|}{0.434}               & $\underline{0.644}$ & 0.451               & \multicolumn{1}{c}{\textbf{0.222}}      \\
DGM        & \textbf{0.790}       & 0.785                & \multicolumn{1}{c|}{\textbf{0.022}}      & 0.601               & 0.521               & \multicolumn{1}{c}{$\underline{0.229}$} \\
GMEL        & 0.249                & \textbf{0.001}       & \multicolumn{1}{c|}{0.995}               & 0.419               & \textbf{0.000}      & \multicolumn{1}{c}{1.000}               \\
NetGAN     & 0.334                & $\underline{0.011}$  & \multicolumn{1}{c|}{0.924}               & 0.422               & 0.029               & \multicolumn{1}{c}{0.978}               \\ \midrule
DiffODGen  & $\underline{0.699}$  & 0.321                & \multicolumn{1}{c|}{$\underline{0.345}$} & \textbf{0.738}      & 0.339               & \multicolumn{1}{c}{0.370}               \\ \bottomrule
\end{tabular}
\vspace{-0.5cm}
\end{table}

\noindent\textbf{Further study of topology similarity and sparsity.}
Next we investigate further the topology similarity and sparsity of generated OD matrices, which is motivated by the observations of NZR$_\text{flow}$ in Table~\ref{tab:topology1}. We check three additional metrics, i.e., accuracy, false negative rate~(predict a real nonzero flow to be zero) and false positive rate~(predict a real zero flow to be nonzero), and report results in Table~\ref{tab:topology2}. Note that accuracy is a biased metric as it does not consider the unbalanced ratio between nonzero-zero flows in real data. Unlike baselines that overestimate either nonzero flows~(GMEL, near-zero FN and near-one FP) or zero flows~(DGM in \textit{New York City}, near-one FN and near-zero FP), the proposed DiffODGen successfully achieves a trade-off, indicating the necessity of learning to generate realistic network topology in city-wide OD matrix generation.

\begin{figure}[t]
\centering
\subfigure[Real flows~(NZR$_\text{flow}$=24.9\%)]{
    \label{nycin}
    \includegraphics[width=0.18\textwidth]{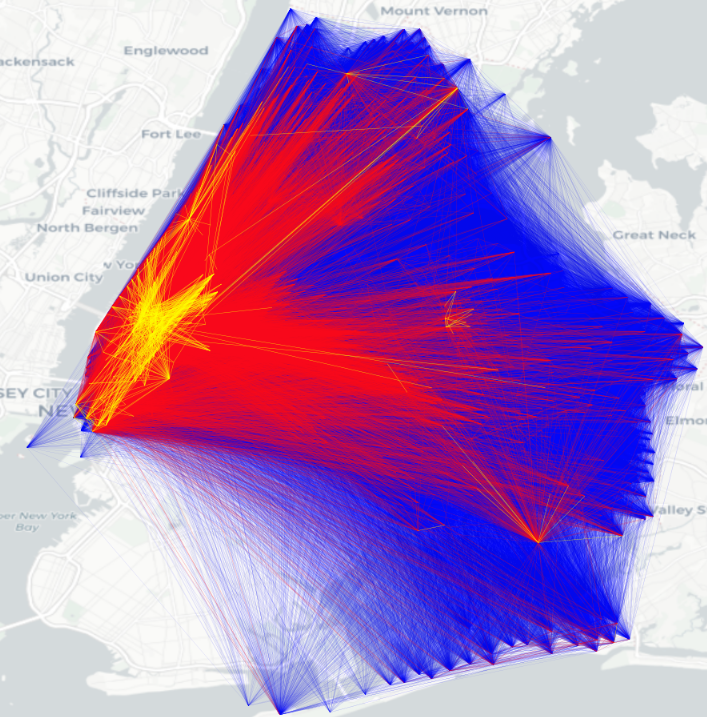}
}
\subfigure[Ours~(NZR$_\text{flow}$=33.0\%)]{
    \label{nycout}
    \includegraphics[width=0.18\textwidth]{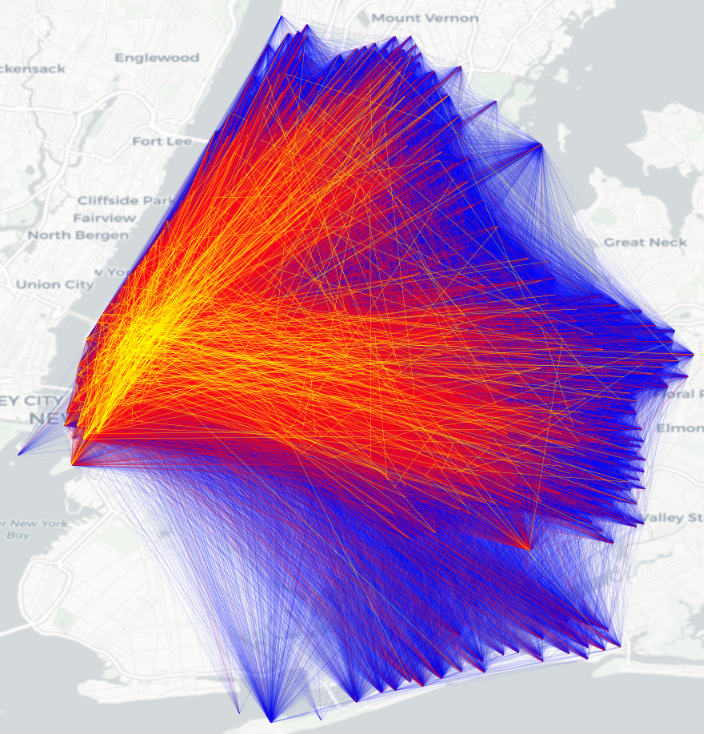}
}
\vfill
\subfigure[DGM~(NZR$_\text{flow}$=6.9\%)]{
    \label{seain}
    \includegraphics[width=0.18\textwidth]{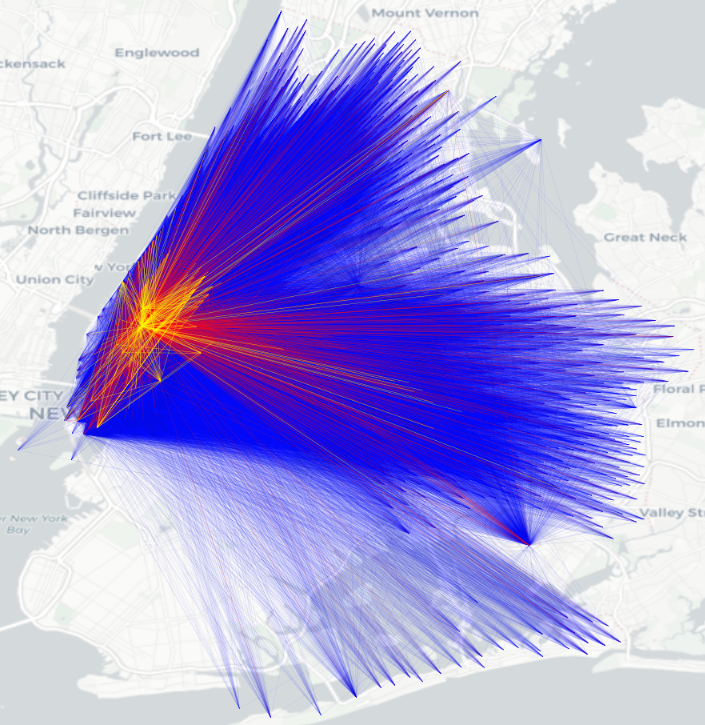}
}
\subfigure[GMEL~(NZR$_\text{flow}$=99.6\%)]{
    \label{seaout}
    \includegraphics[width=0.18\textwidth]{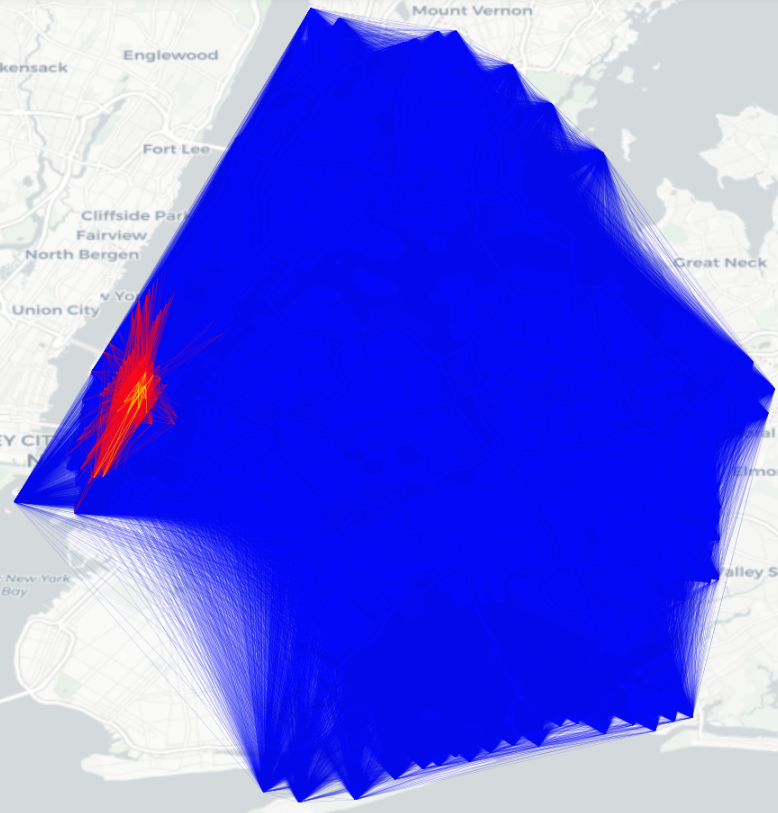}
}
\vspace{-0.5cm}
\caption{Spatial flow visualizations of real OD matrix and generated OD matrices in New York City.}
\vspace{-0.4cm}
\label{fig:viz}
\end{figure}

\noindent\textbf{Flow visualization.}
Finally we visualize the spatial distribution of both real OD flows and generated OD flows in \textit{New York City}~(Figure~\ref{fig:viz}). The results indicate an excellent match to above analysis in Table~\ref{tab:topology1} and Table~\ref{tab:topology2}. Compared to real flows, the flows generated by the proposed DiffODGen is with the highest similarity, while those by DGM and GMEL are either too sparse or too dense.

\begin{figure}[t]
\centering
\subfigure[Flow value error]{
    \label{ablation:err}
    \includegraphics[width=0.25\textwidth]{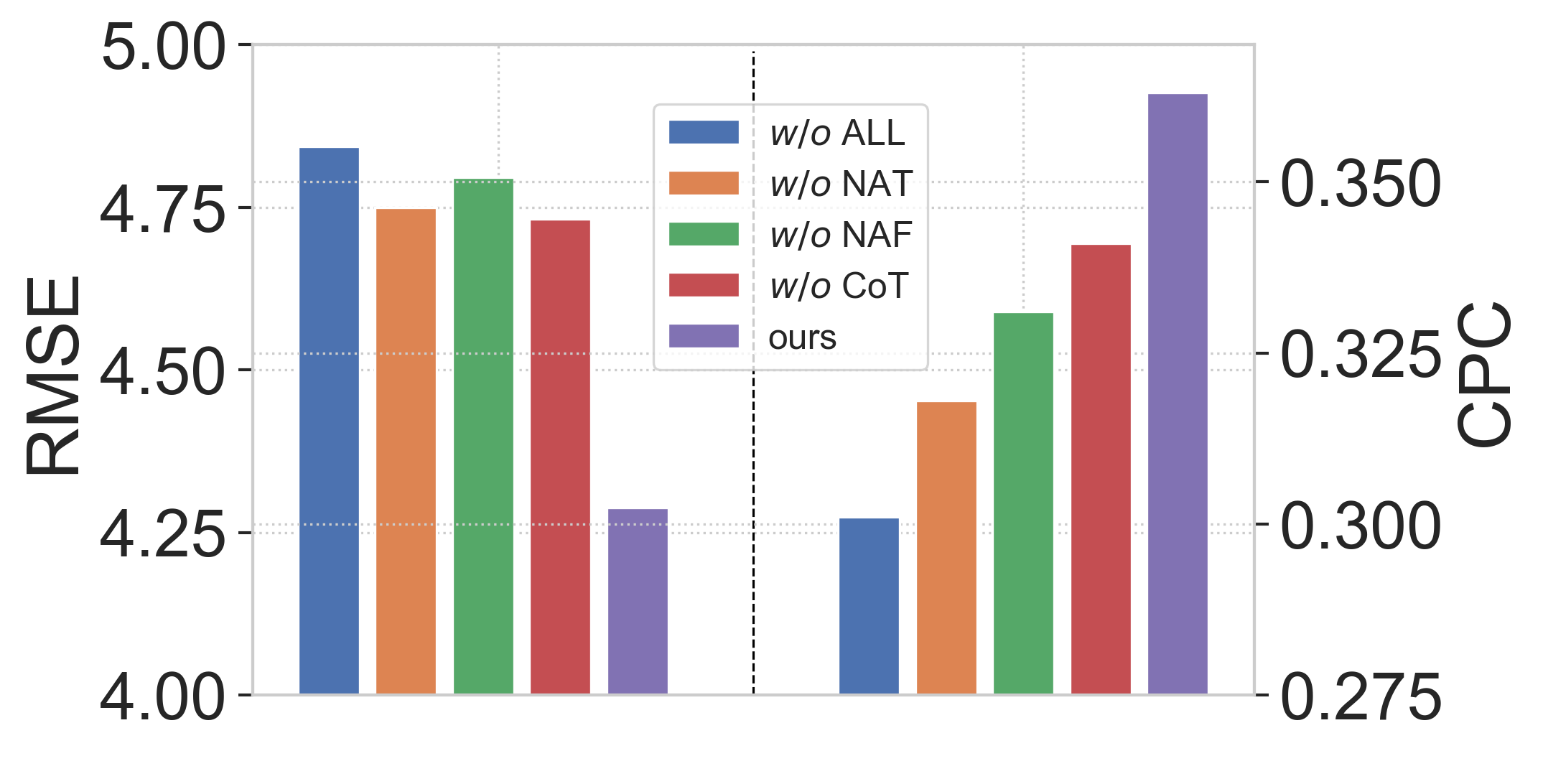}
}
\subfigure[Network statistics similarity]{
    \label{ablation:jsd}
    \includegraphics[width=0.20\textwidth]{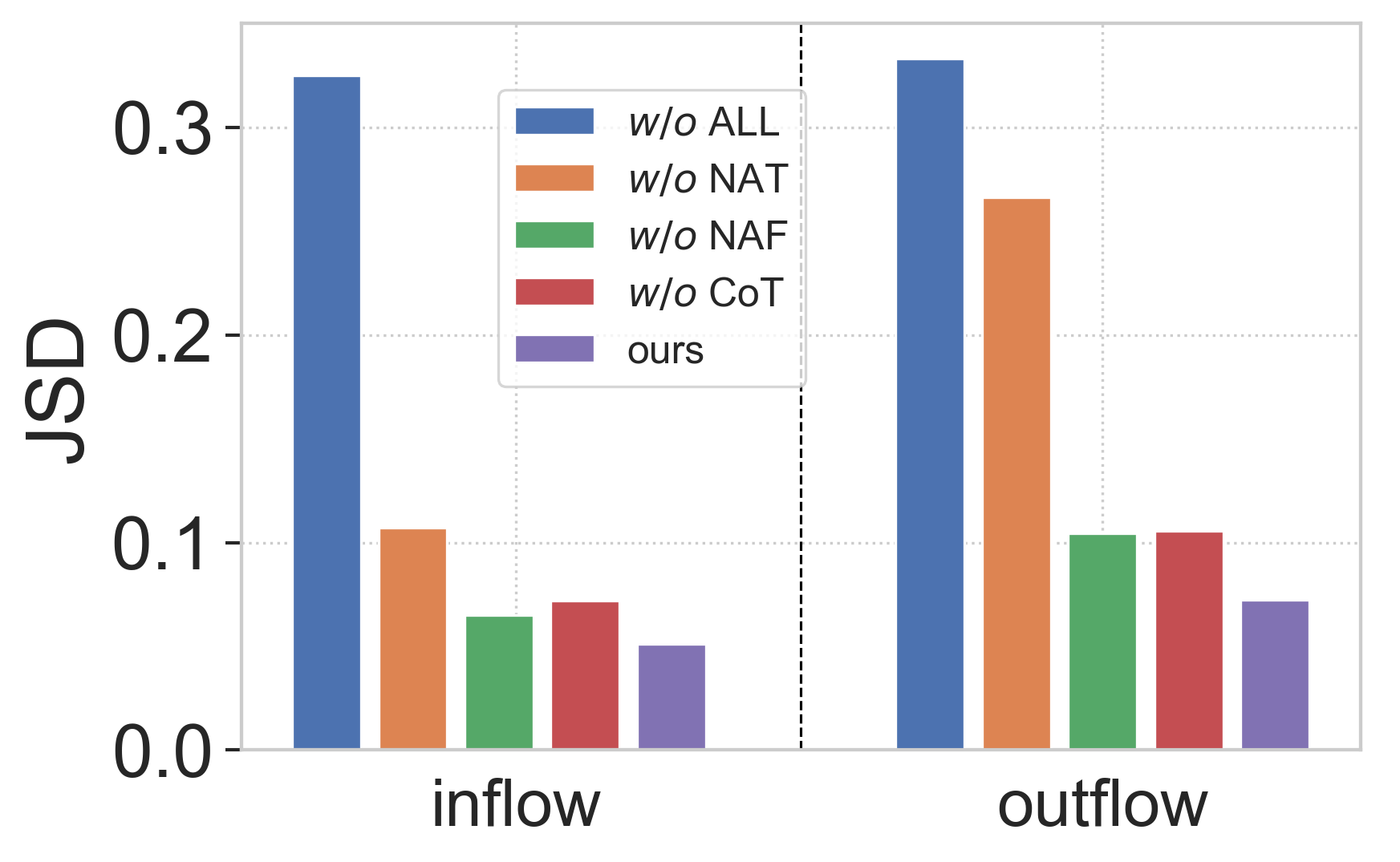}
}
\vspace{-0.5cm}
\caption{Ablation study.}
\vspace{-0.5cm}
\label{fig:ablation}
\end{figure}

\subsection{{Ablation Study}}

\textbf{Node Properties Augmentation for Topology Generation (NAT).} 
We test the effect of node properties augmentation module utilized in the topology diffusion model, and the results show that the accurate and realistic generation of topology structure of mobility flow network is instrumental. From Figure~\ref{ablation:err}, this design will bring up to 14.15\% performance improvement according to CPC, which is the largest in magnitude. From Figure~\ref{ablation:jsd}, the performance of the model on network statistics similarity would also be significantly worse, without this design.

\textbf{Node Properties Augmentation for Topology Generation (NAF).} 
Accordingly, we examine the utility of node properties augmentation applied in the flow diffusion model. The experimental results show that this design also greatly enhances the effect of OD matrix generation with a margin at 9.67\%, according to Figure~\ref{ablation:err}. This shows that the enhancement of graph transformer is useful from both topology and flow volume perspectives. From Figure~\ref{ablation:jsd}, this design also shows comparable behavior in terms of network statistics similarity..

\textbf{Collabrative Training (CoT).} 
By experimenting with just the generation of the topology diffusion model, we found that collaborative training can continue to slightly improve performance based on existing designs according to CPC according to Figure~\ref{fig:ablation}.

\textbf{ALL.} This part of the experiment is intended to examine the total performance improvement of our model designs compared to the direct application of the cascaded graph denoising diffusion model. From the Figure~\ref{fig:ablation}, these model designs are very instrumental. In contrast with naive diffusion models, cascading design can well solve the problem of difficult generation of large-scale OD matrix and improve the effect with a 20.6\% improvement on performance.